\documentclass[sigconf]{acmart}

\AtBeginDocument{%
  }

\usepackage{multirow}
\usepackage{subcaption}
\usepackage{bm}

\copyrightyear{2024}
\acmYear{2024}
\setcopyright{acmlicensed}\acmConference[SA Technical Communications
'24]{SIGGRAPH Asia 2024 Technical Communications}{December 3--6, 2024}{Tokyo,
Japan}
\acmBooktitle{SIGGRAPH Asia 2024 Technical Communications (SA Technical
Communications '24), December 3--6, 2024, Tokyo, Japan}
\acmDOI{10.1145/3681758.3698022}
\acmISBN{979-8-4007-1140-4/24/12}

\setcitestyle{square}

\begin{document}

\title{An Empirical Analysis of GPT-4V’s Performance\\on Fashion Aesthetic Evaluation}

\settopmatter{authorsperrow=4}

\author{Yuki Hirakawa}
\affiliation{%
  \institution{ZOZO Research}
  \city{Tokyo}
  \country{Japan}}
\email{yuki.hirakawa@zozo.com}

\author{Takashi Wada}
\affiliation{%
  \institution{ZOZO Research}
  \city{Tokyo}
  \country{Japan}}
\email{takashi.wada@zozo.com}

\author{Kazuya Morishita}
\affiliation{%
  \institution{ZOZO Research}
  \city{Tokyo}
  \country{Japan}}
\email{kazuya.morishita@zozo.com}

\author{Ryotaro Shimizu}
\affiliation{%
  \institution{ZOZO Research}
  \city{Tokyo}
  \country{Japan}}
\email{ryotaro.shimizu@zozo.com}

\author{Takuya Furusawa}
\affiliation{%
  \institution{ZOZO Research}
  \city{Tokyo}
  \country{Japan}}
\email{takuya.furusawa@zozo.com}

\author{Sai Htaung Kham}
\affiliation{%
  \institution{ZOZO Research}
  \city{Tokyo}
  \country{Japan}}
\email{sai.htaungkham@zozo.com}

\author{Yuki Saito}
\affiliation{%
  \institution{ZOZO Research}
  \city{Tokyo}
  \country{Japan}}
\email{yuki.saito@zozo.com}

\renewcommand{\shortauthors}{Hirakawa et al.}

\begin{abstract}
Fashion aesthetic evaluation is the task of estimating how well the outfits worn by individuals in images suit them. In this work, we examine the zero-shot performance of GPT-4V on this task for the first time. We show that its predictions align fairly well with human judgments on our datasets, and also find that it struggles with ranking outfits in similar colors. The code is available at https://github.com/st-tech/gpt4v-fashion-aesthetic-evaluation.
\end{abstract}

\ccsdesc[100]{Computing methodologies~Scene understanding}

\keywords{vision and language, aesthetic evaluation, fashion}

\begin{teaserfigure}
  \includegraphics[width=\textwidth]{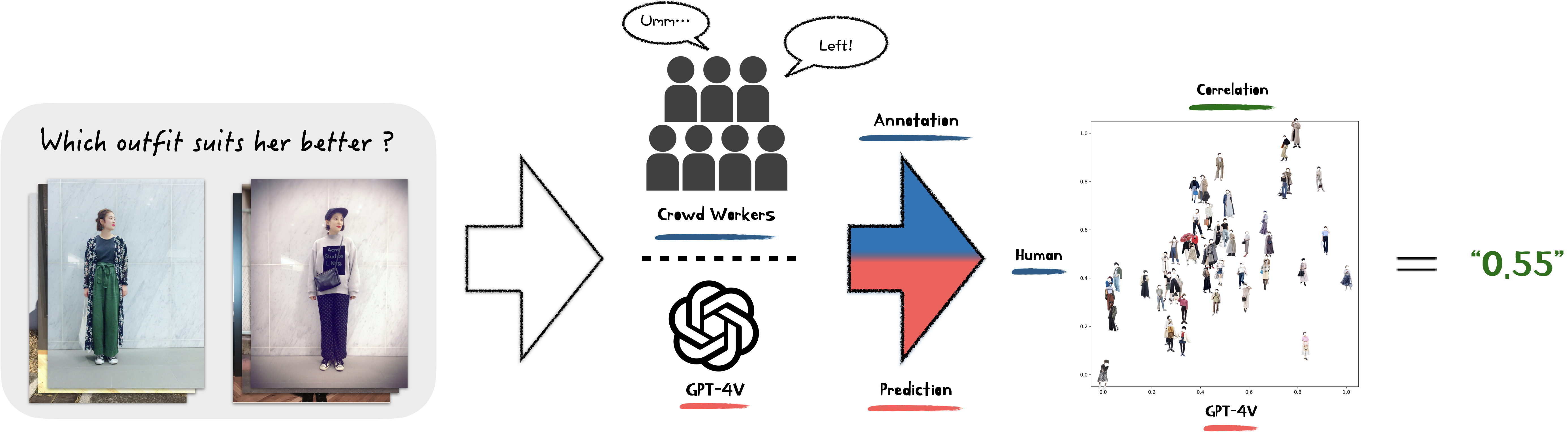}
  \caption{Overview: Evaluation pipeline for assessing GPT-4V's ability to evaluate fashion aesthetics.}
  \Description{Evaluation pipeline for assessing GPT-4V's ability to evaluate fashion aesthetics}
  \label{fig:teaser}
\end{teaserfigure}

\maketitle

\section{Introduction}
Fashion is an effective means of enhancing one’s physical attractiveness ~\citep{article}. However, for those who are not familiar with fashion, it can be challenging to choose the outfits that suit them well. As such, there has been a line of research for many years on fashion aesthetic evaluation, i.e., the task of estimating how well the outfits worn by individuals in images suit them ~\citep{alshan2018learning, neuberger2018learning}. 

To tackle this task, previous work has constructed datasets by asking human annotators to compare pairs of images where people wear different outfits and select the one that looks better ~\citep{alshan2018learning, neuberger2018learning}. Then, they used them to train and evaluate deep learning models that judge fashion aesthetics. However, one limitation is that the sizes of the datasets are limited, making it questionable whether they accurately represent the fashion preferences of the general population. On the other hand, constructing a large-scale dataset is very challenging due to both time and financial cost, which motivates the use of deep learning models to automatically evaluate fashion aesthetics and generate a large number of pseudo labels.

Recently, it has been shown that GPT-4V ~\citep{openai2024gpt4}, which integrates the conversational abilities and world knowledge of large language models (LLMs) ~\citep{brown2020language} with vision foundation models (VFMs) ~\citep{radford2021learning, dosovitskiy2021an}, achieves near human-level performance across various tasks, including image caption generation, visual question answering, and visual commonsense reasoning ~\citep{yang2023dawn}. Refs. ~\citep{zhang2023gpt, wu2023gpteval3d} have also leveraged GPT-4V as an automatic evaluator of multi-modal generative models and demonstrated that it aligns well with human judgments on many tasks.  This suggests that, with well-designed prompts, it might be possible to employ multi-modal LLMs like GPT-4V to automatically assess fashion aesthetics to some degree and aid human annotators in constructing a large-scale dataset for fashion aesthetic evaluation.

In this paper, we make a first attempt to study how well GPT-4V performs on fashion aesthetic evaluation and demonstrate its potential use as an automatic evaluator for this task. Specifically, our contributions are (1) the construction of reliable gold-standard datasets annotated by hundreds of human annotators; and (2) the evaluation of the zero-shot performance of GPT-4V on fashion aesthetic evaluation and how well it aligns with human judgments. 

\begin{figure}[t]
    \centering
    \begin{minipage}[t]{0.8\linewidth}
        \centering
        \includegraphics[width=\linewidth]{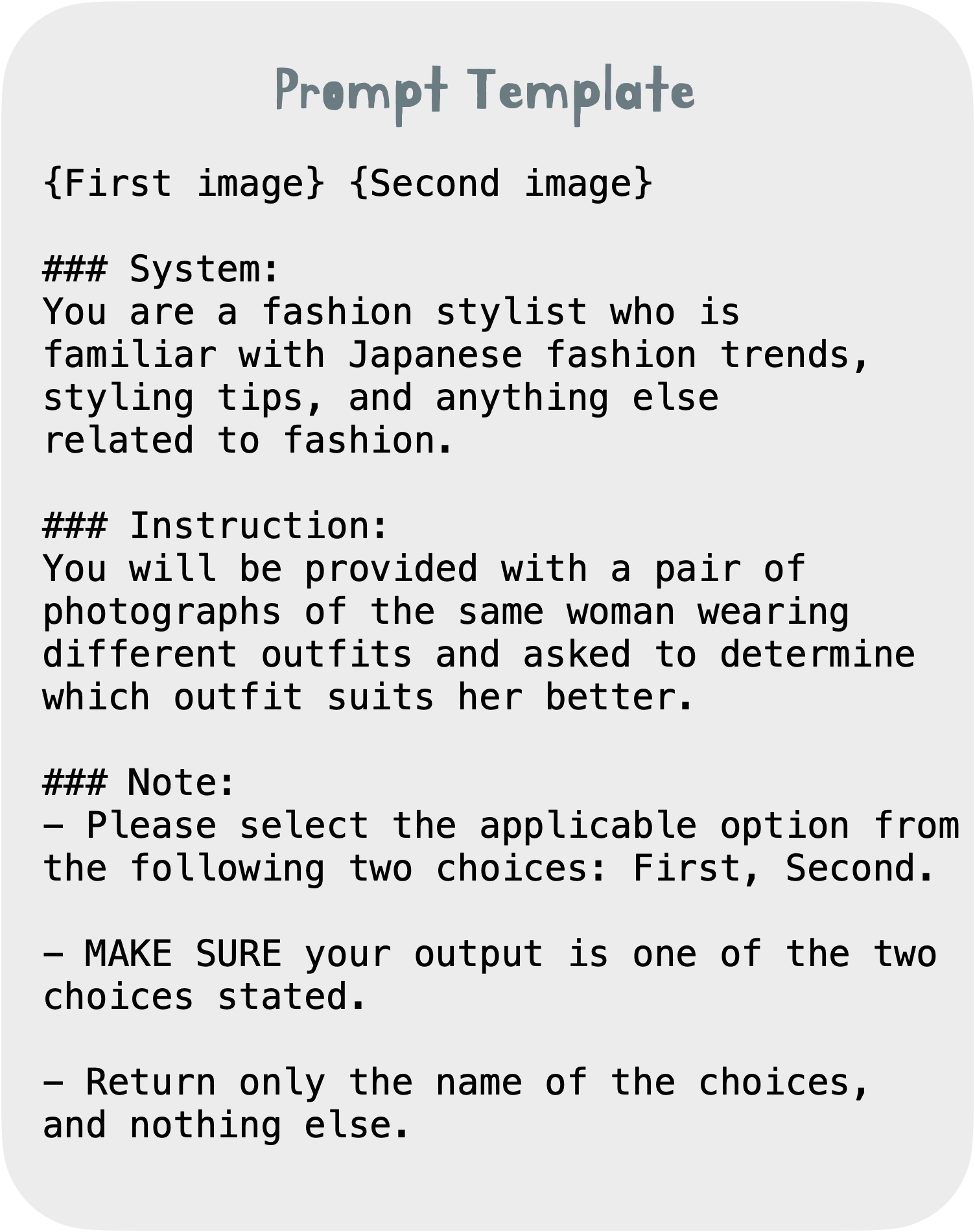}
        \caption{
        The prompt used in our experiments.
        }
        \Description{The prompt used in our experiments.}
        \label{prompt}
    \end{minipage}
    \end{figure}

\begin{table*}[t]
  \caption{Results on Top-K vs Bottom-K Classification.}
  \vspace{-3mm}
  \centering
  \label{tb:tkbk_classification}
  \begin{tabular}{ccccccccccc}
    \toprule
    \multirow{2}{*}{Methods} & \multicolumn{2}{c}{sbj-1}& \multicolumn{2}{c}{sbj-2}&\multicolumn{2}{c}{sbj-3-a}&\multicolumn{2}{c}{sbj-3-b}&\multicolumn{2}{c}{sbj-3-c}\\
    \cmidrule(lr){2-3}\cmidrule(lr){4-5}\cmidrule(lr){6-7}\cmidrule(lr){8-9}\cmidrule(lr){10-11}&$K=10$&$K=50$&$K=10$&$K=50$&$K=10$&$K=50$&$K=10$&$K=50$&$K=10$&$K=50$\\\midrule
    \multicolumn{11}{c}{Accuracy ($\uparrow$)}\\
    \midrule
    View& 0.160 & 0.467 & 0.198 & 0.480 & 0.431 & 0.355 & 0.351 & 0.421 & 0.342 & 0.373\\
    Like & 0.099 & 0.449 & 0.099 & 0.356 & 0.484 & 0.370 & 0.307 & 0.405 & 0.298 & 0.462\\
    Like/View& $\bm{0.593}$ & 0.511 & 0.407 & 0.387 & 0.631 & $\bm{0.629}$ & 0.604 & 0.563 & 0.564 & 0.630\\
    GPT-4V & 0.531 & $\bm{0.584}$ & $\bm{0.531}$ & $\bm{0.618}$ & $\bm{0.638}$ & 0.583 & $\bm{0.738}$ & $\bm{0.631}$ & $\bm{0.582}$ & $\bm{0.670}$\\
    \midrule
    \multicolumn{11}{c}{Calibrated Accuracy ($\uparrow$)}\\
    \midrule
    View& 0.087 & 0.448 & 0.185 & 0.560 & 0.391 & 0.324 & 0.266 & 0.357 & 0.278 & 0.291\\
    Like & 0.130 & 0.458 & 0.074 & 0.493 & 0.435 & 0.305 & 0.266 & 0.374 & 0.216 & 0.388 \\
    Like/View& $\bm{0.739}$ & 0.583 & 0.370 & 0.480 & 0.598 & 0.625 & 0.771 & 0.622 & 0.66 & 0.713\\
    GPT-4V & 0.609 & $\bm{0.698}$ & $\bm{0.593}$ & $\bm{0.853}$ & $\bm{0.837}$ & $\bm{0.691}$ & $\bm{0.991}$ & $\bm{0.857}$ & $\bm{0.691}$ & $\bm{0.869}$\\
    \midrule
    \multicolumn{11}{c}{Conflict Rate ($\downarrow$)}\\
    \midrule
    GPT-4V & 0.716 & 0.573 & 0.741 & 0.658 & 0.578 & 0.552 & 0.298 & 0.634 & 0.538 & 0.538\\
    \bottomrule          
  \end{tabular}
\end{table*}

\section{Dataset Construction}

To assess the performance of GPT-4V on fashion aesthetic evaluation, we constructed reliable gold-standard datasets annotated by hundreds of human annotators. To this end, we first collected snapshots from the largest fashion outfit-sharing app in Japan called WEAR,\footnote{https://wear.jp/} where the users upload their outfit photos. Among all users, we selected three female users (denoted as \textit{sbj-1}, \textit{sbj-2}, and \textit{sbj-3}, resp.) who regularly posted their outfits on this platform. Then, we asked human annotators to judge some of their outfits based on how well the outfits suit them. Specifically, we showed the annotators a pair of snapshots where one of the target users wears different outfits, and asked them which outfits suit her better.\footnote{As a pre-processing step, we applied a background removal tool to each snapshot.} After collecting the pairwise annotations, we created a ranking of the images with corresponding scores for each user; we will explain the details in Section~\ref{section_OpenSkill}.

For sbj-3, we hired 765 female annotators in their 30s via a crowd-sourcing website and asked them to annotate 999 images posted by sbj-3,\footnote{Additionally, we included a dummy dog photo to detect non-serious annotators.} resulting in 132,905 pairwise annotations in total. For sbj-1 and sbj-2, we sampled 30 snapshots posted by each user and collected the annotations from users of our EC platform. Specifically, we asked female users in their 30s whether they were interested in a quick survey to rate outfits (that consists of 30 annotations) for a chance to win points redeemable on our platform. We got 50 and 44 annotators with 1461 and 1268 pairwise annotations for sbj-1 and sbj-2, resp.\footnote{For sbj-1/2/3, we discarded the annotations from non-serious annotators who chose the same option more than 90\% of the time.} We validated the collected annotations by measuring the inter-rater agreement, which we describe in the next section.

\subsection{Annotation Based on OpenSkill}\label{section_OpenSkill}
In previous work, multiple annotators judged the same pair of snapshots to ensure annotation reliability~\citep{alshan2018learning, neuberger2018learning}. However, this approach may not be optimal in terms of scalability, as it reduces the total number of images to be annotated. To mitigate this, we employed the OpenSkill \citep{Joshy2024} algorithm, a Bayesian-inference-based online rating system that is typically used to estimate player skills in competitive games. In our case, each snapshot is treated as a player, and we estimated their scores/ranks based on their match results annotated by humans. This approach is inspired by \citet{hipsterwars}, which employed TrueSkill \citep{herbrich2006trueskill} (an analogous algorithm to OpenSkill) to collect human judgments of fashion styles in snapshots.
Similar to their approach, we selected a pairs of images to match by first choosing the least-matched image and selecting its opponent by sampling an image that is likely to result in a draw based on the draw probability estimated by the tentative OpenSkill scores.

To measure the inter-rater agreement, we randomly split the annotators into two independent groups and calculated Spearman's rank correlation coefficient between the rankings generated by each group. We observed relatively high correlations: 0.716, 0.815, and 0.715 for sbj-1, sbj-2, and sbj-3, indicating that the annotators' perception of fashion aesthetics were more or less similar.\footnote{For sbj-3, we found that some annotators were accidentally assigned to both groups. Therefore, after collecting the annotations, we re-calculated the coefficient by randomly dividing the annotators into two groups and generating rankings using OpenSkill.} To make the most of all the collected annotations, we aggregated the annotations from both groups and re-applied OpenSkill to produce a final ranking, which we use for evaluation throughout our experiments.

\section{GPT-4V as a Fashion Aesthetic Evaluator}

\subsection{Prompt Template} 

We carefully designed a prompt to ensure that GPT-4V understands our task and generates the output in the desired format. As shown in Fig.~\ref{prompt}, we first showed the model a pair of snapshots and prompted it to act as a fashion stylist to elicit its prior fashion knowledge. Our prompt also requires the model to output either ``First'' or ``Second'', where ``First'' indicates that the first outfits suit her better than the other, and vice versa. 
\subsection{Position Bias Calibration} 

When we prompt LLMs to solve multiple-choice questions, their responses can be influenced by the input order of the choices, which is known as the position bias of LLMs ~\citep{wang2023large, zheng2024large}. In our experiments, we also observed that GPT-4V is also subject to this bias, i.e.,\ swapping the input order of the snapshots changes the model's prediction sometimes. Therefore, we prompted the model twice for each pair of snapshots with different orders, and if the predictions were inconsistent, we rendered it inconclusive and selected the winner at random. 

\section{Experiments}
\begin{table}[t]
  \caption{Spearman’s rank correlation on the ranking task. }
  \vspace{-3mm}
  \centering
  \label{tb:end2endsimulation}
    \begin{tabular}{cccccc}
        \toprule
        Methods & sbj-1 & sbj-2 & sbj-3-a & sbj-3-b  & sbj-3-c\\
        \midrule
        View& 0.083 & 0.208 & -0.140 & -0.269 & -0.359\\
        Like & $\bm{0.181}$ & 0.155 & -0.193 & -0.332 & -0.282\\
        Like/View& 0.047 & -0.055 & 0.111 & 0.137 & 0.270\\
        gpt-4o & 0.117 & $\bm{0.364}$ & $\bm{0.360}$ & $\bm{0.519}$ & $\bm{0.472}$\\\midrule
        Human* & 0.716 & 0.815 & 0.791 & 0.711 & 0.759\\
        \bottomrule
    \end{tabular}
\end{table}


\subsection{Baselines}
We compare GPT-4V\footnote{We use ``gpt-4o-2024-05-13'' and set the temperature to 0.0 in our experiments.} against three baseline models: View, Like, and Like/View models, all of which are based on user feedback and insights on the WEAR app. Specifically, View and Like sort images based on how many times the snapshots were viewed or liked on the app, and Like/View sorts them based on the number of likes per view. We assume that if the outfits suit the person well, the snapshot would be viewed and liked by many users.

\begin{figure*}[t]
    \centering
    \begin{minipage}[t]{0.32\linewidth}
        \centering
        \includegraphics[width=\linewidth]{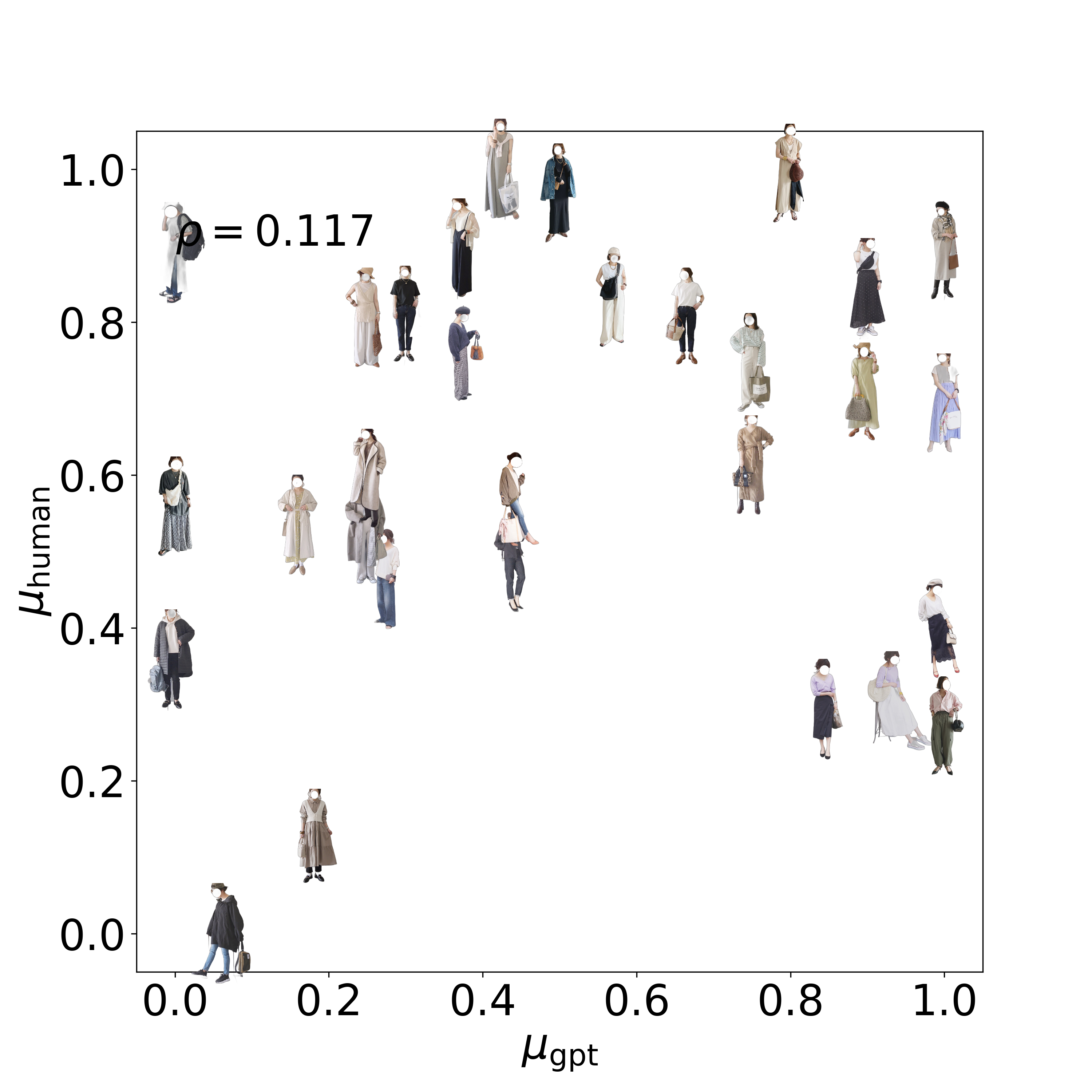}
        \subcaption{sbj-1}
        \label{yo}
    \end{minipage}
    \hfill
    \begin{minipage}[t]{0.32\linewidth}
        \centering
        \includegraphics[width=\linewidth]{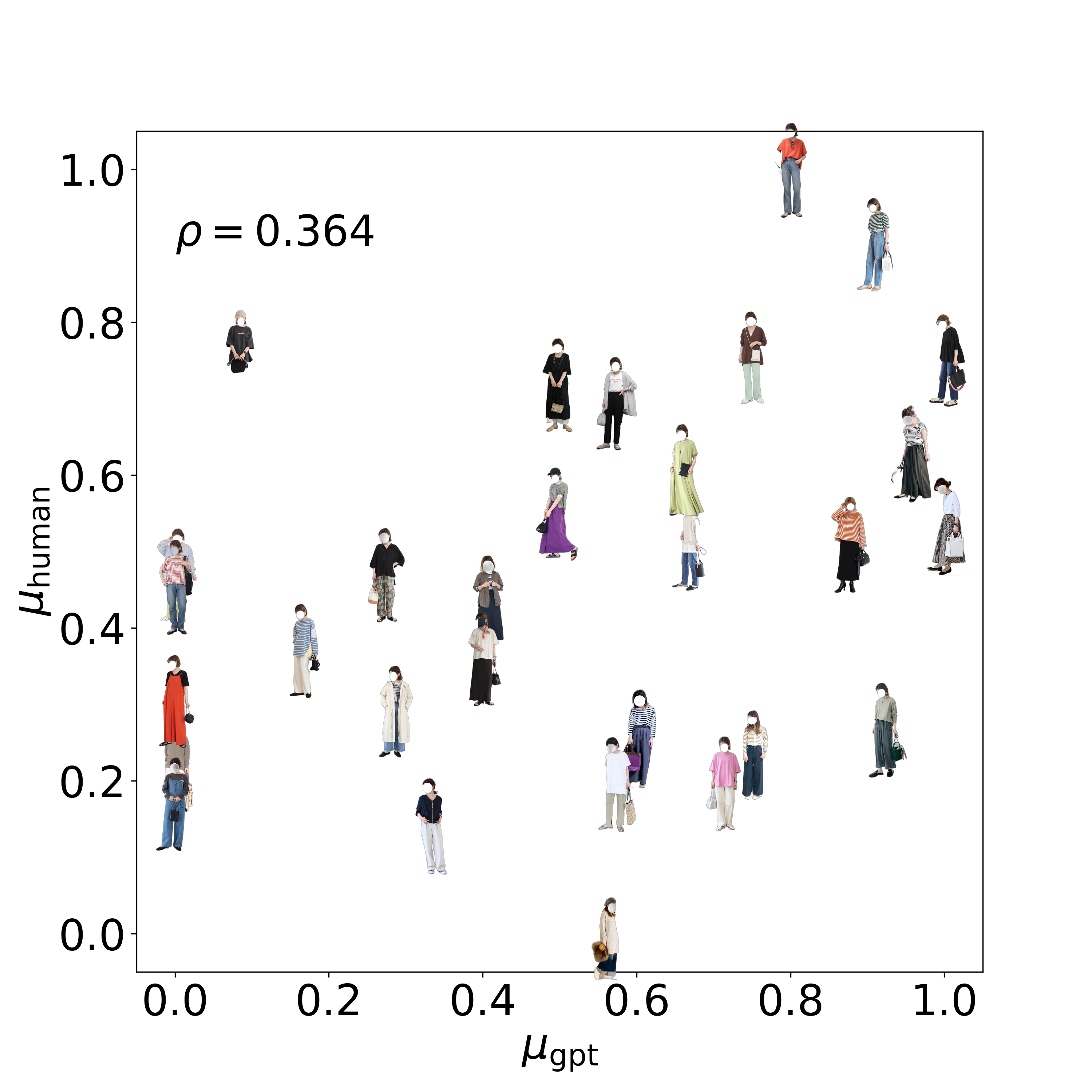}
        \subcaption{sbj-2}
        \label{kana}
    \end{minipage}
    \hfill
    \begin{minipage}[t]{0.32\linewidth}
        \centering
        \includegraphics[width=\linewidth]{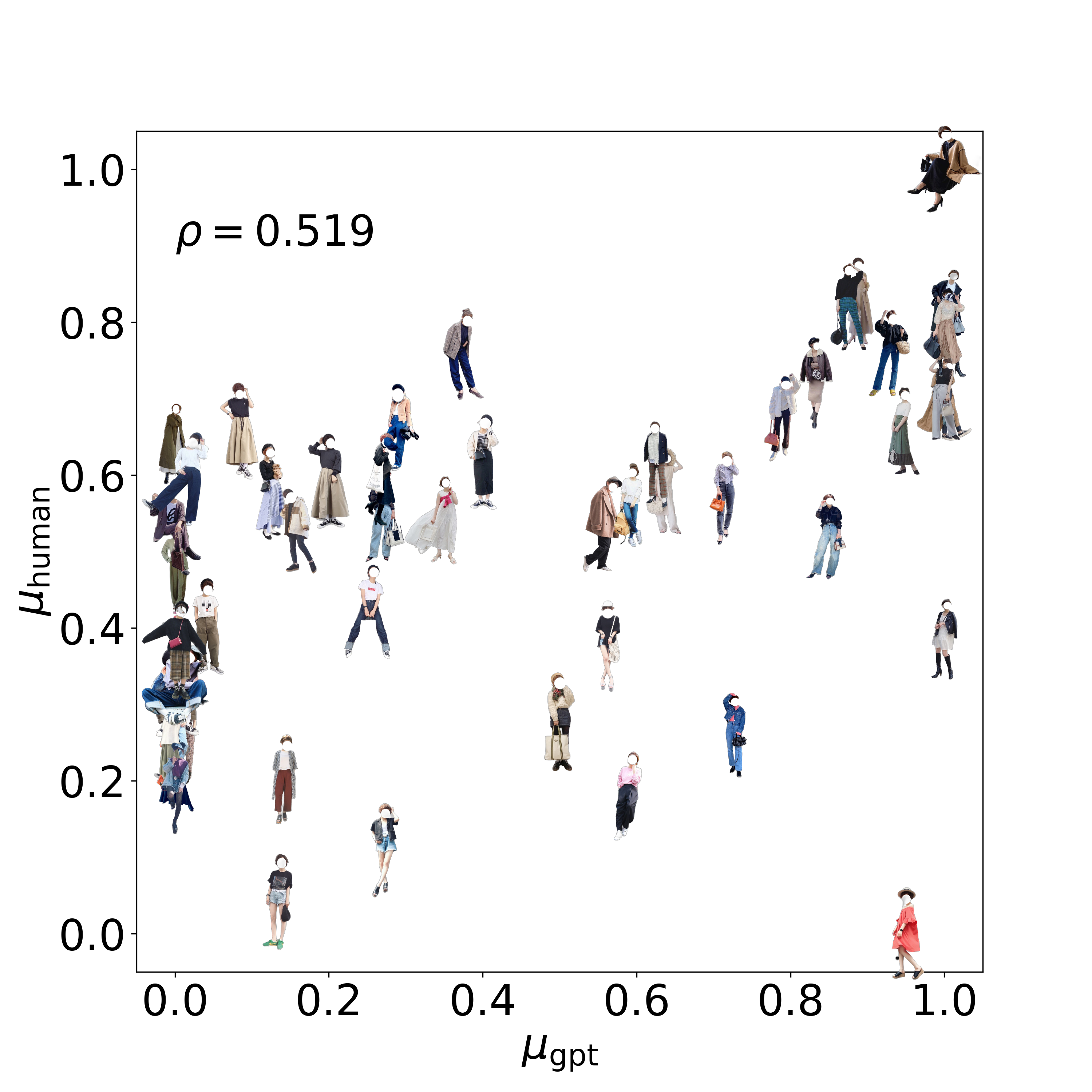}
        \subcaption{sbj-3-b}
        \label{yuki(group2)}
    \end{minipage}
    \Description{The x-axis corresponds to scores predicted by GPT-4V and y-axis to OpenSkill scores generated by human annotations.}
    \caption{The x-axis corresponds to scores predicted by GPT-4V and y-axis to OpenSkill scores generated by human annotations.}
    \label{fig:gptvshuman}
\end{figure*}
\subsection{Evaluation}
Given a list of images ranked by humans, we evaluate models on two tasks: image classification and ranking. For classification, we first sample a pair of images, each from the top-$K\%$ and bottom-$K\%$ of the ranking. We evaluate the baselines based on whether they rank the one from the top-$K\%$ subset better than the other. For GPT-4V, we feed a pair of images and ask it which one is better using the prompt described in Fig.~\ref{prompt}.  We calculate the classification accuracy with $K \in \{10, 50\}$. For ranking, we calculate Spearman’s rank correlation coefficient ~\citep{spearman1961proof} between the ground-truth and predicted rankings. To generate rankings using GPT-4V, we perform pairwise comparisons for all possible combinations of the images (twice for each pair with different input orders to mitigate the position bias) and rank them based on the win ratio. 

For evaluation, we use the whole datasets of sbj-1/2 and also three subsets of sbj-3 (denoted as sbj-3-a/b/c, resp.), each of which consists of 50 images randomly sampled from the whole data.

\subsection{Results}

Table \ref{tb:tkbk_classification} presents the results of the image classification. The first four rows denote the classification accuracy of each model, indicating that GPT-4V performs better than the chance level prediction (0.500) on all datasets and also surpasses the reasonable baselines overall. We also look at the accuracy on the subsets of the instances for which GPT-4V makes consistent judgments regardless of the order of the input images; the results are shown below \textit{Calibrated Accuracy}. Note that \textit{Conflict Rate} denotes the proportion of the instances affected by the position bias (the lower, the better). They indicate that the performance of GPT-4V increases substantially, demonstrating its potential as an evaluator of fashion aesthetics.

Table \ref{tb:end2endsimulation} shows Spearman’s rank correlation coefficients between the rankings generated by the models and humans. The last row denotes the correlation between the rankings generated by annotators from different groups during the data construction process. While those inter-rater correlations are not exactly comparable to the others since we aggregated all annotations from both groups to create a final ranking, they serve as a good estimator of the upper bound on this task. The results show that GPT-4V performs fairly well and outperforms the baselines except for sbj-1. 

As a qualitative analysis, Fig.~\ref{fig:gptvshuman} shows the scatter plots of the images where the x and y axes denote the scores generated by GPT-4V and humans, respectively. The leftmost plot shows the results on sbj-1; compared to the other datasets, it clearly lacks variety in color, suggesting that GPT-4V struggles with ranking outfits in similar colors.
\section{Limitations}
One limitation of this work is the lack of diversity in our datasets, as they focus on the snapshots of three Japanese women in different outfits. Likewise, we obtained annotations from women in their 30s in Japan, suggesting that the datasets are biased towards Japanese female fashion. Therefore, it is important to include a wider range of demographics and styles and see whether our findings generally hold true.   Furthermore, the prompt we used for GPT-4V is rather simple, and we may get better results by using well-known prompting techniques such as few-shot and chain-of-thought prompting.
\section{Conclusion}
In this work, we explored the potential use of GPT-4V as a fashion aesthetic evaluator. To assess its capacities, we constructed reliable gold-standard datasets annotated by hundreds of human evaluators using the OpenSkill rating system. 
We found that GPT-4V’s predictions align fairly well with human judgments in the image classification tasks across all of our datasets, as well as in the ranking tasks, except for a dataset composed of outfits in similar colors.
Our findings suggest that GPT-4V holds promise as a tool for fashion evaluation, although further refinement may be needed to address specific challenges.
\begin{acks}
We thank Takahiro Kawashima for his useful comments. 
\end{acks}
\bibliographystyle{ACM-Reference-Format}
\bibliography{sample-base}

\end{document}